\pdfoutput=1

\documentclass[11pt]{article}

\usepackage[]{ACL2023}

\usepackage{times}
\usepackage{latexsym}

\usepackage[T1]{fontenc}

\usepackage[utf8]{inputenc}

\usepackage{microtype}

\usepackage{inconsolata}

\usepackage[pdftex]{graphicx}
\usepackage{bm}
\usepackage{amssymb}
\usepackage{amsmath}
\usepackage{hyperref}
\usepackage{adjustbox}
\usepackage{booktabs}
\usepackage{multirow}
\usepackage{hyperref}
\usepackage{textcomp}
\usepackage{todonotes}

\usepackage[normalem]{ulem}
\useunder{\uline}{\ul}{}

\usepackage{array}
\newcommand{\PreserveBackslash}[1]{\let\temp=\\#1\let\\=\temp}
\newcolumntype{C}[1]{>{\PreserveBackslash\centering}p{#1}}
\newcolumntype{R}[1]{>{\PreserveBackslash\raggedleft}p{#1}}
\newcolumntype{L}[1]{>{\PreserveBackslash\raggedright}p{#1}}

\title{RECAP: Retrieval-Enhanced Context-Aware Prefix Encoder for Personalized Dialogue Response Generation}

\author{Shuai Liu \qquad Hyundong J. Cho \qquad Marjorie Freedman \\ {\bf Xuezhe Ma} \qquad {\bf Jonathan May} \\
        Information Sciences Institute \\ University of Southern California \\ \texttt{\{liushuai, jcho, mrf, xuezhema, jonmay\}@isi.edu}}

\begin{document}
\maketitle
\begin{abstract}
Endowing chatbots with a consistent persona is essential to an engaging conversation, yet it remains an unresolved challenge.
In this work, we propose a new retrieval-enhanced approach for personalized response generation.
Specifically, we design a hierarchical transformer retriever trained on dialogue domain data to perform personalized retrieval and a context-aware prefix encoder that fuses the retrieved information to the decoder more effectively.
Extensive experiments on a real-world dataset demonstrate the effectiveness of our model at generating more fluent and personalized responses. 
We quantitatively evaluate our model's performance under a suite of human and automatic metrics and find it to be superior compared to state-of-the-art baselines on English Reddit conversations.\footnote{Our code and data are publicly available at \url{https://github.com/isi-nlp/RECAP}.}

\end{abstract}

\section{Introduction}

As tremendous successes have been achieved on open-domain dialogue generation \cite{zhang-etal-2020-dialogpt, cho-may-2020-grounding, roller-etal-2021-recipes, https://doi.org/10.48550/arxiv.2208.03188}, personalized dialogue models have started to draw attention because of their ability to generate consistent and engaging conversations and their potential time-saving utility in on-message predictive generation \cite{wu-etal-2021-personalized, 10.1145/3404835.3462828, zhong-etal-2022-less}.
To generate persona-consistent responses, these models condition on not only dialogue context but user personas, which can be either explicitly given or implicitly learned from the user conversations.
Early works mostly focus on modeling explicit personas \cite{zhang-etal-2018-personalizing, DBLP:journals/corr/abs-1901-09672, ijcai2019p721, song-etal-2021-bob}.
These methods rely on dialogue data paired with user traits, profiles or persona description sentences, which are difficult to collect in practice.
Moreover, explicit personas usually only contain a few user traits (e.g. age, gender, and location) or a few profile sentences, so the amount of information carried with them is limited, which restricts the models' capability to capture and then express more nuanced personalization.
Later works develop methods for automatically extracting personas \cite{mazare-etal-2018-training, wu-etal-2020-getting}, in order to help improve content diversity, as compared to that seen in explicit personas. 
However, these extraction methods still cannot fully use all information from user history conversations.

Recent works address these issues by incorporating user dialogue history as their implicit profiles \cite{wu-etal-2021-personalized, 10.1145/3404835.3462828, zhong-etal-2022-less}.
These methods generate personalized responses in two phases, retrieving relevant conversations from the user history and fusing the retrieved information to the generator. 
In the first phase, these methods retrieve a subset of conversations from a user's conversation history \cite{10.1145/3404835.3462828, zhong-etal-2022-less}.
In the second phase, the retrieved conversations are fused into a decoder by manipulating output logits \cite{10.1145/3404835.3462828, wu-etal-2021-personalized} or by adding prompt tokens to the input \cite{zhong-etal-2022-less}.
Even though the implicit profile approach is shown to be the most robust and scalable among all approaches on real-world datasets, it still has some potential weaknesses.
Approaches to the retrieval phase include using recent conversations \cite{10.1145/3404835.3462828} or using conversations based on current context similarity, according to an out-of-domain model \cite{wu-etal-2021-personalized, zhong-etal-2022-less}. These approaches have the potential to lose important personal information, which may lead to unexpected behavior and poorly motivated retrieval. In the fusion phase, neither output logit manipulation nor input token prompting may fully leverage the capability of the pre-trained decoder.
In this work, we focus on the implicit user profile approach, but specifically address the weaknesses in both retrieval and fusion phases.


We present RECAP, a \textbf{R}etrieval-\textbf{E}nhanced \textbf{C}ontext-\textbf{A}ware \textbf{P}refix encoder for personalized dialogue response generation.
Similar to other implicit user profile methods, our model is based on a retrieval-fusion approach, which first retrieves persona-relevant information, and then fuses it with conversation context at decode-time.
Unlike previous work, which does not take the purpose of the retrieval task into consideration, our hierarchical transtormer retriever is trained specifically to retrieve information that will best communicate a user's persona.
Unlike previous work, which approaches fusion by concatenating retrieved information at the input level \cite{zhong-etal-2022-less} or manipulating logits at the output level \cite{10.1145/3404835.3462828, wu-etal-2021-personalized}, we adopt a continuous pre-layer prefix approach \cite{li-liang-2021-prefix, liu-etal-2022-p}, along with a two-step cross-attention projection \cite{DBLP:conf/iclr/HumeauSLW20, DBLP:conf/nips/MaKWZMMZ21}, both of which have been shown to be beneficial. These novelties result in a better personalized dialogue model.

Our main contributions in this work are:
\begin{itemize}
  \item We design a hierarchical transformer retriever that can perform personalized history retrieval based on different target users using their history conversations.
  \item We design a context-aware prefix encoder that can encode context-relevant information from user histories and fuse the information to the generator effectively through the prefix.
  \item The two modules combined achieve state-of-the-art performance on personalized dialogue response generation for English Reddit conversations by generating fluent and personalized responses for unseen users, as shown in automatic and human evaluations.
\end{itemize}

\section{Methodology}
In this section, we formalize the personalized dialogue response generation task and introduce our proposed RECAP method.

\subsection{Task Definition}
Our goal is to build a personalized dialogue model that generates persona-consistent responses with a target user's \textit{history} of conversations.
Formally, we have a set of users $\mathcal{U}$ but will henceforth assume user $u \in \mathcal{U}$ is the \textit{target user}, that is, the user we wish to personalize, User $u$'s history is represented as a set of context-response pairs $\mathcal{H}_{u} = \{(\textbf{c}_1^{u}, \textbf{r}_1^{u}), \cdots, (\textbf{c}_T^{u}, \textbf{r}_T^{u})\}$, where a \textit{context} $\textbf{c}_t^{u}$ is a sequence of one or more turns that starts at the beginning of a conversation and ends with the turn immediately before the single turn \textit{response} $\textbf{r}_t^{u}$, which is by definition authored by $u$.\footnote{Some, but not all, of the turns in $\textbf{c}_t^{u}$ may have been authored by $u$.} Given some \textit{current} (context, response) pair $(\textbf{c}^{u}, \textbf{r}^{u}) \notin  \mathcal{H}_{u}$, we seek to maximize

\begin{equation}\label{gen-obj}
    p(\textbf{r}^{u}|\textbf{c}^{u}, \mathcal{H}_u) =
    \prod_{i=1}^{|\textbf{r}^u|}
    p(r^u_i | \textbf{c}^u, \textbf{r}^u_{<i}, \mathcal{H}_u)
\end{equation}
where $\textbf{r}^u_{<i}$ represents tokens preceding token $r^u_i$ in $\textbf{r}^u$.\footnote{Unless otherwise noted, we drop the superscript henceforth.}


\subsection{Model Overview}
RECAP consists of two main modules: a retrieval module (RE), which selects user history responses, and a context-aware prefix encoder (CAP), which converts the selected responses into a suitable dense prefix. The prefix, when prepended to our transformer decoder's intermediate states as in \citet{liu-etal-2022-p}, yields personalized generation. In the following we describe each of RE and CAP in more detail.

\subsection{Retrieval Module (RE)}
\label{sec:re}

\begin{figure}[t!]
    \centering 
    \includegraphics[width=1.0\columnwidth]{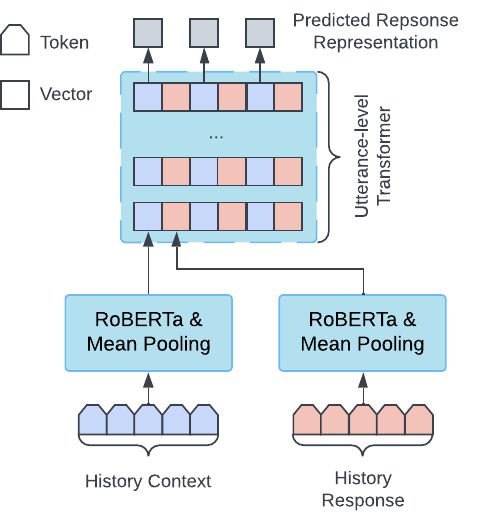}
    \caption{The architecture overview of the retrieval module (RE) based on hierarchical transformer.}
    \label{fig:re_overview}
    \vspace{-0.4cm}
\end{figure}

We follow a standard bi-encoder retrieval approach as in \citet{Wu_Fisch_Chopra_Adams_Bordes_Weston_2018}: a dense representation is formed for each of $u$'s candidate dialogue turns, which we regard as \textit{documents}, as well as for a \textit{query} representing the context of the conversation for which a turn will be generated. The set of documents that is closest (as measured by cosine) to the query is returned.

Using such retrieval methods for dialogue personalization is not novel \cite{isbell2006cobot, zhang-etal-2018-personalizing, wu-etal-2021-personalized, 10.1145/3404835.3462828, zhong-etal-2022-less}, but previous work simply either retrieves the user's most recent turns \cite{10.1145/3404835.3462828} or simply queries for turns based on the similarity to the current conversation or a predicted topic \cite{zhong-etal-2022-less}. Our retrieval model, by contrast, forms a query based on an \textit{a priori} predicted next turn given the current context and user history.

The representation of the predicted next turn is learned based on not only the context of the immediate conversation, but also based on every other conversation known to involve $u$, forming an ersatz persona. This can present a problem. Prolific users can potentially have had a long history of  conversations from which to choose to represent  an implicit persona, and it is unrealistic to use this entire history when generating a single response. 
Furthermore, existing works have shown that using a small subset of the history is more beneficial to implicit persona-based generation, as doing so reduces noise and computation time  \cite{wu-etal-2021-personalized, zhong-etal-2022-less}. 
Inspired by the hierarchical dialogue model \cite{DBLP:journals/corr/SerbanSBCP15, Serban_Sordoni_Bengio_Courville_Pineau_2016} and the hierarchical transformer \cite{DBLP:journals/corr/abs-1910-10781, zhang-etal-2019-hibert}, we build a response prediction model that  takes as input a user's available history and the current dialogue context in an efficient manner. The hierarchical architecture is shown in \autoref{fig:re_overview}.

Specifically, we first concatenate all turns in the oldest history context, $\textbf{c}_1$, encode them with a pre-trained RoBERTa model \cite{DBLP:journals/corr/abs-1907-11692}, and form a fixed-length representation from the mean of the last hidden states corresponding to each token. To this we add positional embedding $\textbf{p}_1$ indicating it is the first history context known, and utterance type embedding $\textbf{y}_c$ indicating a context representation. We then do the same for the analogous $\textbf{r}_1$, and in turn for all other context and response pairs in the history. In general, the inputs to the next level of the hierarchical transformer are formed as follows: 
\begin{align}
    \textbf{e}_{\textbf{c}_t} &= \mathrm{mean}(\mathrm{RoBERTa}(\textbf{c}_t)) + \textbf{p}_t + \textbf{y}_c \\
    \textbf{e}_{\textbf{r}_t} &= \mathrm{mean}(\mathrm{RoBERTa}(\textbf{r}_t)) + \textbf{p}_t + \textbf{y}_r
\end{align}
We term these \textit{utterance-level} embeddings and pass $[\textbf{e}_{\textbf{c}_1}, \textbf{e}_{\textbf{r}_1}, \cdots, \textbf{e}_{\textbf{c}_T}, \textbf{e}_{\textbf{r}_T}]$ to an utterance-level transformer, yielding the sequence of hidden representations $[\textbf{h}_{\textbf{c}_1}, \textbf{h}_{\textbf{r}_1}, \cdots, \textbf{h}_{\textbf{c}_T}, \textbf{h}_{\textbf{r}_T}]$. 
We train the transformer to predict the ground truth response representation and minimize cosine similarity between the predicted and ground truth representation, i.e.
\begin{equation}
    \mathcal{L} = \sum^T_{t=1} 1 - \frac{\textbf{h}_{\textbf{r}_t}\cdot\textbf{g}_{\textbf{r}_t}}{\lvert\textbf{h}_{\textbf{r}_t}\rvert\lvert\textbf{g}_{\textbf{r}_t}\rvert}
\end{equation}
where $\textbf{g}_{\textbf{r}_t}$ is the ground-truth representation of the response at time $t$ encoded by an off-the-shelf sentence transformer. 
A causal mask is applied during training to prevent attention to future utterances.

This general architecture can be specialized by changing the underlying pretrained RoBERTa model used for token-level embedding. In particular, we consider two types of representation: (1) a \textit{style} representation, which we obtain by encoding histories with an off-the-shelf content-independent style representation model \cite{wegmann-etal-2022-author},\footnote{\url{https://huggingface.co/AnnaWegmann/Style-Embedding}} and (2) a representation encoded by a sentence transformer \cite{reimers-gurevych-2019-sentence}\footnote{\url{https://huggingface.co/sentence-transformers/all-distilroberta-v1}} which we term a \textit{semantic} representation, to contrast the style representation.

For retrieval, we first predict the style and semantic representations of the response to be generated and retrieve the appropriately embedded history responses whose style/semantic representations are the most similar to the predicted style/semantic representations. The retrieved history responses are then passed to the context-aware prefix encoder (Section \ref{sec:cap}) for further encoding.

\subsection{Context-Aware Prefix Encoder (CAP)}
\label{sec:cap}

\begin{figure*}[t!]
    \centering 
    \includegraphics[width=2.0\columnwidth]{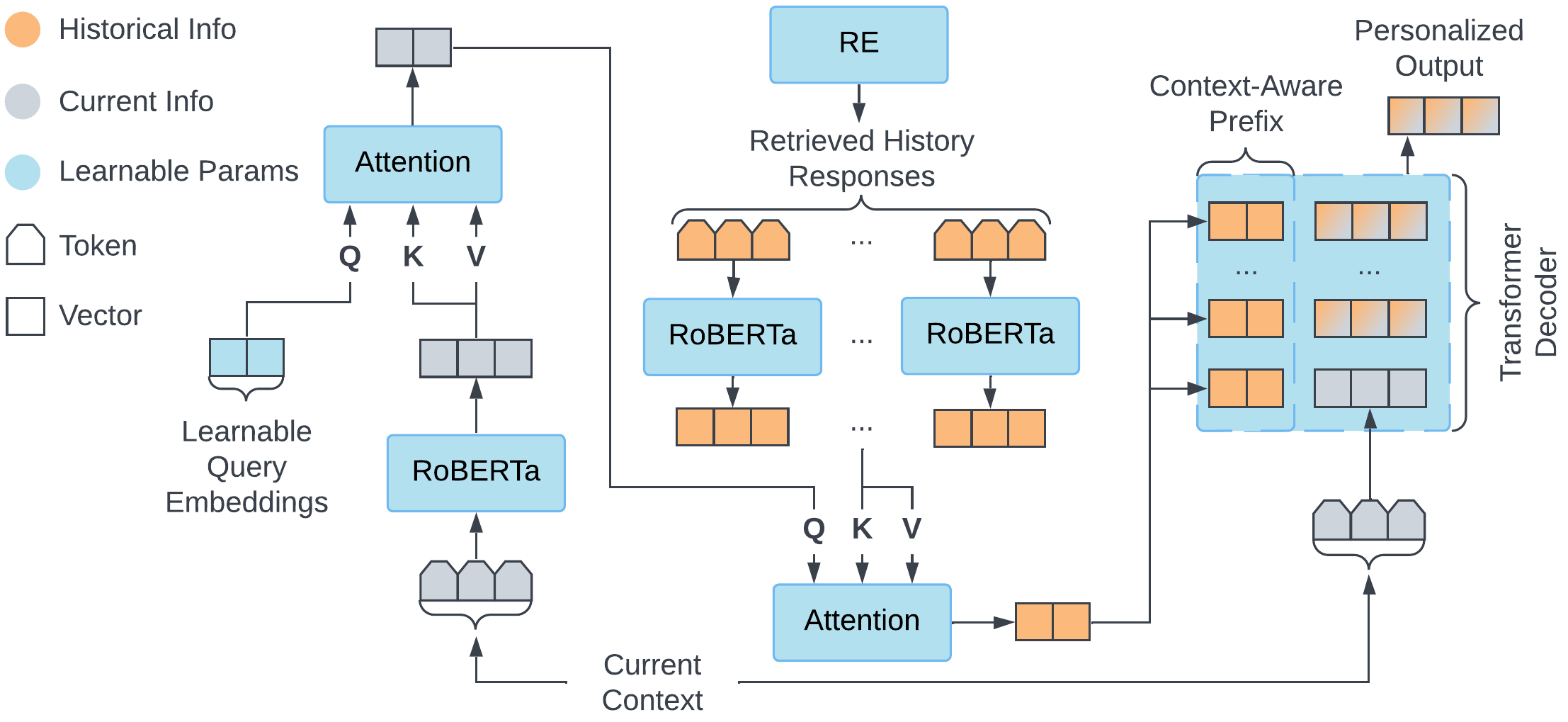}
    \caption{The architecture overview of the context-aware prefix encoder (CAP) and the decoder generator.}
    \label{fig:cap_overview}
    \vspace{-0.4cm}
\end{figure*}


The purpose of the CAP module is to  project the history responses  retrieved by RE (Section \ref{sec:re}) onto a fixed-length prefix vector. This vector is then prepended to the transformer decoder hidden states as a prefix.
The architecture is illustrated in \autoref{fig:cap_overview}. 
CAP first encodes the current dialogue context and each of the retrieved responses to continuous representations with a pre-trained RoBERTa encoder \cite{DBLP:journals/corr/abs-1907-11692}.\footnote{\url{https://huggingface.co/docs/transformers/model_doc/roberta}} We use the same method as \citet{liu-etal-2021-three} to add positional embeddings to user history response representations:
\begin{align}
    \textbf{C} &= \mathrm{RoBERTa}(\textbf{c}) \\
    \textbf{H}_i &= \mathrm{RoBERTa}(\textbf{h}_i) + \textbf{q}_i
\end{align}

\noindent where $\textbf{c}$ is the current context, $\textbf{h}_i$ is the $i$-{th} retrieved history response, $\textbf{C}$ and $\textbf{H}_i$ are the last hidden states from the application of RoBERTa to the every token position of $\textbf{c}$ and the $i$-{th} retrieved history response, respectively, and $\textbf{q}_i$ is a history positional embedding for retrieved history $i$.
All $\textbf{H}_i$'s are then concatenated to a long vector sequence $\textbf{H} = [\textbf{H}_1;\cdots;\textbf{H}_{t-1}]$.

Inspired by the cross-attention context projection operation  \cite{DBLP:conf/iclr/HumeauSLW20, DBLP:conf/nips/MaKWZMMZ21}, CAP projects the long vector sequence $\textbf{H}$ onto a short fixed-length prefix with two cross-attention operations, which we denote as $\mathrm{Attn}(\textbf{Q}, \textbf{K}, \textbf{V})$ to indicate which information source is used as query, key, and value, respectively.\footnote{The information source for key and value is always the same, as is typical.} Separate query, key, and value matrices are learned for each of the two operations.

The first cross-attention operation queries $\textbf{C}$ with learnable query embeddings $\textbf{E} \in \mathbb{R}^{Nd}$ and projects to fixed-length representation $\textbf{P}_c \in \mathbb{R}^{Nd}$; $N$ is a chosen hyperparameter and $d$ is RoBERTa's token embedding dimension:
\begin{equation}
    \textbf{P}_c = \mathrm{Attn}(\textbf{E}, \textbf{C}, \textbf{C})
\end{equation}
Then, the second operation queries the user history representations $\textbf{H}$ with $\textbf{P}_c$ to obtain the fixed-length context-aware user history representations $\textbf{P}_h \in \mathbb{R}^{Nd}$:
\begin{equation}
    \textbf{P}_h = \mathrm{Attn}(\textbf{P}_c, \textbf{H}, \textbf{H})
\end{equation}
Finally, similar to a memory vector projection \cite{li-etal-2020-optimus}, $\textbf{P}_h$ is projected onto $\mathbb{R}^{LNd}$ with a linear layer and then separated into $L$ $d$-dimensional vector sequences with length $N$, corresponding to the $L$ layers in the transformer decoder. Each of these sequences is then prepended to the transformer decoder hidden state in the analogous layer.

\subsection{Generator}
We use the pre-trained DialoGPT \cite{zhang-etal-2020-dialogpt}\footnote{\url{https://huggingface.co/docs/transformers/model_doc/dialogpt}} as the generator. 
Personalized information is fused to the generation process through the prefix vectors encoded by CAP, as described in Section \ref{sec:cap}. 
We further train the parameters in DialoGPT together with the CAP module to maximize the (log of the) objective in \autoref{gen-obj}.

\section{Experiments}

\subsection{Dataset}
We extract a personalized conversation dataset from Reddit on pushshift.io \cite{DBLP:conf/icwsm/BaumgartnerZKSB20}.\footnote{\url{https://files.pushshift.io/reddit/}} We choose only from the conversations from August 2019 to June 2021 to avoid test data leakage when using pre-trained models. Each sample in the dataset consists of three entries: user name, context (i.e preceding turns in the conversation), and response. Since the total number of samples is very large, we randomly select 115,000 users. For each selected user, we keep only the 10 most recent samples to train the generator and the 100 most recent samples as history conversations to retrieve from. Unlike existing works \cite{wu-etal-2021-personalized, 10.1145/3404835.3462828, zhong-etal-2022-less} that use the same users for training, validation, and test, we partition the dataset by user. In this way, we can test the model's ability to generalize to unknown users. Specifically, we select 100,000, 5,000, and 10,000 distinct users for each of training, validation, and test, respectively. 


\subsection{Baseline Models}
We compare our model with four baseline models including the state-of-the-art personalized dialogue model.
\begin{itemize}
  \item \textbf{DialoGPT:} A large-scale pre-trained dialogue response generation model trained on Reddit conversations \cite{zhang-etal-2020-dialogpt}.
  \item \textbf{DialoGPT w/ history responses:} We directly prepend retrieved history responses to the DialoGPT input (i.e. dialogue context).
  \item \textbf{DHAP:\footnote{\url{https://github.com/zhengyima/DHAP}}} A model that generates personalized responses by building a dynamic context-aware user profile representation from user history conversations and then employing a personalized decoder with a copy mechanism \cite{10.1145/3404835.3462828}. 
  We enhance DHAP with pre-trained transformers for fair comparison to our model.
  \item \textbf{MSP:\footnote{\url{https://github.com/bangbangbang12315/MSP}}} The state-of-the-art personalized dialogue model. MSP generates personalized responses by prepending selected tokens directly to the DialoGPT input. 
  The tokens are selected by a three-step hierarchical refiner \cite{zhong-etal-2022-less}.
\end{itemize}

\subsection{Implementation Details}
Our implementation is based on HuggingFace's Transformers \cite{wolf-etal-2020-transformers}\footnote{\url{https://huggingface.co/docs/transformers}} and Sentence Transformer \cite{reimers-gurevych-2019-sentence}\footnote{\url{https://www.sbert.net}} codebases.
We experiment with different settings and hyperparameters; the ones that work the best are discussed below.
We initialize all encoders from the pre-trained RoBERTa-base model \cite{DBLP:journals/corr/abs-1907-11692}\footnote{We specifically use RoBERTa rather than some other core representation for compatibility purposes; RoBERTa and DialoGPT share vocabulary, which is required by two of the baseline models, DHAP and MSP.}  and initialize all decoders from the pre-trained DialoGPT-small model \cite{zhang-etal-2020-dialogpt}.
RoBERTa's embedding dimension, $d$, is 768, and $N$, the prefix length, is set to 30 to align with the prompt length used in MSP.
The two projection attentions in CAP are both single-head attentions. 
The number of history responses are 10, and for the models without a  retrieval module (DialoGPT + history, DHAP, and CAP), the 10 most recent history responses are used. 
The utterance-level transformer has 768 hidden dimension, six layers, and a 12-head self-attention in each layer. 
We train all models using the AdamW optimizer \cite{DBLP:conf/iclr/LoshchilovH19} with learning rate 5e-5 and linear learning rate schedule for 10 epochs. 
The best models are chosen based on the validation perplexity.
For generation, we use nucleus (top-p) sampling \cite{Holtzman2020The} with $p=0.8$.

\begin{table}[!ht]
\begin{adjustbox}{max width=\columnwidth}
\begin{tabular}{lcc}
\toprule
                   & \# Params & Training Time (hr) \\ \midrule
DialoGPT           & 124M      & 8                  \\
DialoGPT + history & 124M      & 37                 \\
DHAP               & 431M      & 25                 \\
MSP                & 437M      & 15                 \\ \midrule
RE                 & 198M      & 13                 \\
CAP                & 269M      & 22                 \\ \bottomrule
\end{tabular}
\end{adjustbox}
\caption{Total number of parameters and training time (on 2 \texttimes\ A40 GPUs) for all neural models.}
\label{table:computation}
\end{table}

All models including the baseline models are trained on 2 \texttimes\ A40 GPUs with half precision training. The total number of parameters and training time are shown in \autoref{table:computation}. For MSP, the numbers are sums over all sub-modules (i.e. three refiners and DialoGPT generator). CAP includes both the context-aware prefix encoder and the DialoGPT generator.

\subsection{Evaluation Metrics}

\subsubsection{Automatic Evaluation}

In this section, we discuss the automatic evaluation metrics we use to evaluate all models. We group them into four categories. 
The first two categories measure the general performance of the models, while the last two measure the personalization ability.

\noindent \textbf{Perplexity and Token-overlap Metrics} \quad We first evaluate the performance of our model with several of the most commonly used automatic evaluation metrics for dialogue response generation, including perplexity, BLEU-1 and -2 \cite{papineni-etal-2002-bleu}, ROUGE-L \cite{lin-och-2004-automatic}, and METEOR \cite{banerjee-lavie-2005-meteor}. 
Perplexity evaluates how well a language model predicts the sample.
Lower perplexity means the model can generate a more fluent response and generalizes better  \cite{blei2003latent}. 
The  other metrics are word or n-gram overlap metrics with a reference utterance.
A higher score means a higher similarity between the generated text and the ground-truth text since they have more words or phrases in common. 

\noindent \textbf{Learning-based Metrics} \quad Many learning-based metrics backed with pre-trained models have been developed.
They are shown to be more robust and correlate better with human judgement than token overlap metrics, though issues have been raised regarding their inherent biases \cite{gowda-etal-2021-macro-average}.
In this work, we select two of the most popular learning-based metrics:  BERTScore \cite{bert-score} and BLEURT \cite{sellam-etal-2020-bleurt}.
These two methods also measure how similar a candidate response is to the reference response; higher scores mean higher similarity.

\noindent \textbf{Style Metrics} \quad To measure the models' ability to capture personal writing styles, we employ a pre-trained style representation model \cite{wegmann-etal-2022-author} for evaluation.
We form two metrics based on the style model: (1) embedding similarity and (2) contrastive authorship verification (CAV) accuracy.
Embedding similarity is simply the cosine similarity between the style embedding of the generated response and that of the ground-truth response.
For CAV accuracy, we construct a domain-controlled \cite{wegmann-etal-2022-author} dataset with response triplets built from a generated response anchor and a pair of positive/negative ground-truth responses.
The positive example and negative example are from the same author as the anchor and a randomly sampled author, respectively.
With domain control, we only choose the negative example from the same subreddit as the anchor and the positive example from a different subreddit.
To evaluate, we calculate the percentage of the triplets for which the style model judges the generated anchor response to be more similar to the positive ground-truth response than to the (randomly sampled) negative example.
For both style metrics, a higher score indicates better personalization.


\begin{table}[!ht]
\centering
\begin{adjustbox}{max width=\columnwidth}
\begin{tabular}{lR{5em}R{5em}R{4em}}
\toprule
       & Train & Validation & Test \\
\midrule
Age    & 1378  & 344        & 585  \\
Gender & 1823  & 455        & 784  \\
MBTI   & 5542  & 1385       & 2131 \\
\bottomrule
\end{tabular}
\end{adjustbox}
\caption{Number of train, validation, and test users in the Pandora dataset.}
\label{table:pandora}
\end{table}

\noindent \textbf{Personal Traits Metrics} \quad A good personalized response model should also be able to reflect the personal traits of the target user. 
Therefore, a personal traits classifier is also used as a evaluation method in previous works.
\citet{DBLP:journals/corr/abs-1901-09672} evaluate their model on traits of  age, gender, and location, while \citet{xing-fernandez-2018-automatic} proposed a evaluation method based on a personality classifier.
In this work, we select three personal traits for evaluation: age, gender,\footnote{Due to data limitation issues, we simplify gender identity as a binary.} and Myers-Briggs Type Indicator (MBTI) \cite{briggs1995gifts}.\footnote{We acknowledge there is extensive criticism of MBTI in the psychology community \cite{capraro2002myers,pittenger2005cautionary} and do not address the validity of fixed personality types, of MBTI as an approach to determining types, or even of the general predictability of MBTI labels given dialogue text in this work. However, following other work \cite{kishima-etal-2021-construction,sang-etal-2022-mbti}, we include this metric, which indicates the degree to which the generated responses for a user are as predictive of self-declared MBTI labels as the user's actual responses. The MBTI metric is not the only metric we use, and its inclusion is meant to accompany the others to indicate a consistent trend toward user-like generation.}
We train a model for each trait on the PANDORA dataset \cite{gjurkovic-etal-2021-pandora},\footnote{\url{https://psy.takelab.fer.hr/datasets/all/pandora}} then attempt to determine traits based on generated responses. A good personalized model should generate output that allows a trait classifier to guess traits about as well as it can when given actual responses.

For age, we train a linear regression model and report the Pearson correlation coefficient \cite{benesty2009pearson} between the predicted age and the ground-truth age.
For gender and the four MBTI categories, we train a logistic regression model for each and report the classification F1 score.
For all three metrics, higher scores are better.
We select users for personal traits evaluation from the PANDORA dataset, which is independent from our Reddit test set.
\autoref{table:pandora} shows the statistics for the PANDORA dataset.

The input features, for all models, are the most frequent 40,000 TF-IDF weighted 1-3 word ngrams.
The logistic regression models use the L-BFGS solver.
The best C value (i.e. inverse of regularization strength) for gender and the four MBTI categories (i.e. I/E, S/N, T/F, J/P) are 10, 1, 10, 50, and 0.1, respectively.
All models can be trained within 20 seconds on a single CPU with scikit-learn \cite{scikit-learn}.\footnote{\url{https://scikit-learn.org/stable}}

\subsubsection{Manual Evaluation}
We also conduct a manual evaluation. 
We randomly sample 100 examples from the test set for all models and hire two well-educated volunteer annotators (one of the authors and a friend of one of the authors). 
The annotators evaluate responses on three criteria: fluency, coherency, and persona consistency.
All criteria are scaled from 1 to 3 (from disagree to strongly agree).
First, for fluency, we only show the response and ask ``is the response overall readable and fluent?''
Then, for coherency, we also show the preceding turns in the conversation and ask ``does the response serve as a valid continuation of the preceding conversation?''
Finally, for persona consistency, we show five ground-truth responses written by the target author and ask ``does the response seem like it would have been written by the author of the given texts?''

\section{Results}

In this section, we discuss the experimental results and further analysis. 
Due to limited time and computational resources, we only report the results from a single run, and run statistical significance tests.\footnote{Please refer to \autoref{sec:ttest} for details.}

\begin{table*}[]
\centering
\begin{adjustbox}{max width=\textwidth}
\begin{tabular}{@{\extracolsep{\fill}}lcccccccccc@{}}
\toprule
\multicolumn{1}{c}{\multirow{4}{*}{Model}} & \multicolumn{5}{c}{Reddit}                                                           & \multicolumn{3}{c}{PANDORA}                         \\ \cmidrule(rl){2-6} \cmidrule(rl){7-9}
\multicolumn{1}{c}{} &                & Token-overlap  & Learning-based  & \multicolumn{2}{c}{Style Metric} & \multicolumn{2}{c}{Demographic}   & MBTI            \\ \cmidrule(rl){3-3} \cmidrule(rl){4-4} \cmidrule(rl){5-6} \cmidrule(rl){7-8} \cmidrule(rl){9-9}
\multicolumn{1}{c}{}                       & PPL\textdownarrow            & ROUGE-L\textuparrow        & BLEURT\textuparrow          & Embed Sim\textuparrow       & CAV Acc\textuparrow        & Age\textuparrow             & Gender\textuparrow          & Average\textuparrow         \\ \midrule
DialoGPT                                   & 31.25$^\ddagger$          & 8.49$^\ddagger$           & 0.2421$^\ddagger$          & 22.15$^\ddagger$           & 51.14$^\ddagger$          & 0.0425$^\ddagger$          & 0.6103$^\ddagger$          & 0.4992$^\ddagger$          \\
DialoGPT + history                         & 29.66$^\ddagger$          & 9.84$^\ddagger$           & 0.2482$^\ddagger$          & 40.66$^\ddagger$           & 64.02          & 0.1008$^\ddagger$          & 0.6763$^\ddagger$          & 0.5130$^\ddagger$          \\
DHAP                                       & 29.99$^\ddagger$          & 9.73$^\ddagger$           & 0.2511$^\ddagger$          & 37.14$^\ddagger$           & 61.53$^\ddagger$          & 0.0829$^\ddagger$          & 0.6653$^\ddagger$          & 0.5119$^\ddagger$          \\
MSP                                        & 30.47$^\ddagger$          & 9.36$^\ddagger$           & 0.2453$^\ddagger$          & 34.72$^\ddagger$           & 59.61$^\ddagger$          & 0.1226$^\ddagger$          & 0.6816$^\ddagger$          & 0.5019$^\ddagger$          \\ \midrule
CAP                                        & \textbf{29.44} & 10.09$^\ddagger$          & 0.2534$^\ddagger$          & 40.38$^\ddagger$           & 63.71$^\ddagger$          & \textbf{0.2051} & {\ul 0.7077}$^\dagger$    & 0.5167$^\dagger$          \\
RECAP-style                                & 29.54$^\ddagger$          & 10.05$^\ddagger$          & 0.2525$^\ddagger$          & \textbf{41.40}  & {\ul 64.14}    & {\ul 0.1822}    & 0.7001$^\dagger$          & {\ul 0.5265}    \\
RECAP-semantic                             & 29.50$^\ddagger$          & \textbf{10.33} & \textbf{0.2749} & 39.65$^\ddagger$           & 64.00          & 0.1699$^\dagger$          & \textbf{0.7303} & \textbf{0.5276} \\
RECAP-mixed                                & {\ul 29.47}$^\dagger$    & {\ul 10.27}$^\dagger$    & {\ul 0.2557}$^\dagger$    & {\ul 40.74}$^\dagger$     & \textbf{64.17} & 0.1392$^\ddagger$          & 0.6962$^\dagger$          & 0.5242          \\ \midrule \midrule
Ground-truth                               & -              & -              & -               & -               & 66.20          & 0.2617          & 0.7477          & 0.5257          \\ \bottomrule
\end{tabular}
\end{adjustbox}
\caption{The automatic evaluation results on Reddit and PANDORA datasets with selected metrics.The best and second best results in each column are shown in \textbf{bold} and {\ul underline}, respectively. Scores for ground-truth are not available for metrics calculated based on ground-truth, and they are shown by "-". "$\dagger$" and "$\ddagger$" indicates statistically significant difference for $p<0.05$, between the best or the top two models, respectively, determined by t-test.}
\label{table:selected-results}
\end{table*}

\begin{table}[!ht]
\centering
\begin{adjustbox}{max width=\columnwidth}
\begin{tabular}{lC{4em}C{4.9em}C{4em}}
\toprule
\multicolumn{1}{c}{Model} & Fluency\textuparrow       & Coherency\textuparrow     & Persona\textuparrow       \\ \midrule
DialoGPT                  & 2.75$^\ddagger$          & 2.27$^\ddagger$          & 1.58$^\ddagger$          \\
DialoGPT + history        & 2.77          & 2.28$^\ddagger$          & 1.84$^\ddagger$          \\
DHAP                      & 2.72$^\ddagger$          & 2.28$^\dagger$          & 1.76$^\ddagger$          \\
MSP                       & 2.73$^\ddagger$          & 2.29$^\ddagger$          & 1.85$^\ddagger$          \\ \midrule
CAP                       & 2.72$^\ddagger$          & 2.31          & 1.90$^\ddagger$          \\
RECAP-style               & 2.77          & 2.28$^\ddagger$          & \textbf{2.03} \\
RECAP-semantic            & \textbf{2.80} & \textbf{2.35} & 1.92$^\ddagger$          \\
RECAP-mixed               & {\ul 2.79}    & {\ul 2.33}    & {\ul 2.00}    \\ \midrule \midrule
Ground-truth              & 2.84          & 2.40          & 2.47          \\ \bottomrule
\end{tabular}
\end{adjustbox}
\caption{The human evaluation results on the Reddit dataset. The best and second best results in each column are shown in \textbf{bold} and {\ul underline}. "$\dagger$" and "$\ddagger$" indicates statistically significant difference for $p<0.05$, between the best or the top two models, respectively, determined by t-test.}
\label{table:manual-results}
\end{table}

\subsection{Automatic Evaluation Results}

\autoref{table:selected-results} shows the automatic evaluation results for all models on selected metrics. 
For conciseness, we only show a representative or aggregated metric for similar metrics, but the full results are shown in \autoref{sec:more} and are generally consistent with the representative results shown in \autoref{table:selected-results}.
In nearly all cases, the top two results in all automatic metrics are from our models.
Without the retriever, the CAP model already outperforms the baseline models on most automatic metrics.
With the retrieval enhancement, the RECAP models achieve better scores on most automatic metrics.
Specifically, with style retrieval enhancement, the RECAP model obtains better style embedding similarity, CAV accuracy, and average MBTI F1 score, which indicates  better performance at reflecting the target author's writing style.
With semantic retrieval enhancement, the RECAP model achieves the best scores on token-overlap metrics and learning-based metrics, which indicates it can generate responses that are more similar to the ground-truth.
Moreover, combining two enhancement methods by mixing half retrieved history responses from each retriever also combines the strength of the two RECAP models.
Even though the combination also weakens the improvements, we can still see that the RECAP-mixed model is at least the second best on all metrics on the Reddit dataset.


\subsection{Human Evaluation Results}

\autoref{table:manual-results} shows the human evaluation results. 
Cohen's $\kappa$ and Krippendorff’s $\alpha$ between the two annotators are $\kappa=0.617$ and $\alpha=0.687$, respectively.
The $\kappa$ shows a substantial agreement between the two annotators, and the $\alpha$ indicates that a tentative conclusion could be drawn from the human evaluation results \cite{antoine-etal-2014-weighted}.
Even though there are some minor inconsistencies, both human annotation results and automatic evaluation results on the Reddit dataset agree on the top two models, which are RECAP-semantic and RECAP-mixed for general response quality, and RECAP-style and RECAP-mixed for style/persona metrics.
Furthermore, RECAP-mixed is the overall second best model under human evaluation.


\subsection{Style Consistency Analysis}

Even though the automatic and human metrics give us a general idea of the model performance, these scores are not very interpretable. 
To further understand style consistency beyond the metric scores, we conduct a case analysis similar to \citet{wegmann-etal-2022-author} by inspecting whether the models can capture some aspects of writing style.
Specifically, we select three aspects mentioned by \citeauthor{wegmann-etal-2022-author}: last punctuation (i.e. whether the response ends with a punctuation mark), contraction spelling (i.e. whether the response uses ``n't'' or ``nt'' in contractions like ``didn't''), and casing (i.e. whether the response is all lowercased).
For each aspect, we calculate the percentage of generated responses that match the ground-truth style.
\autoref{table:style-consistency} shows the results, which indicate that most of our models can capture all three selected aspects more effectively than the baseline models. The lone exception is the RECAP-style model which is slightly worse than the DialoGPT + history model on last punctuation and casing aspects.

\begin{table}[]
\begin{adjustbox}{max width=\columnwidth}
\begin{tabular}{lC{4em}C{4em}C{4em}}
\toprule
\multicolumn{1}{c}{Model} & Punc.\textuparrow           & Cont.\textuparrow           & Casing\textuparrow          \\ \midrule
DialoGPT                  & 0.3538          & 0.3822          & 0.3300          \\
DialoGPT + history        & 0.4117          & 0.4333          & 0.4180          \\
DHAP                      & 0.3829          & 0.4121          & 0.3831          \\
MSP                       & 0.3795          & 0.4064          & 0.3788          \\ \midrule
CAP                       & 0.4144          & 0.4415          & 0.4183          \\
RECAP-style               & 0.4112          & 0.4403          & 0.4172          \\
RECAP-semantic            & \textbf{0.4232} & \textbf{0.4520} & \textbf{0.4248} \\
RECAP-mixed               & {\ul 0.4178}    & {\ul 0.4451}    & {\ul 0.4195}    \\ \bottomrule
\end{tabular}
\end{adjustbox}
\caption{Style consistency analysis results for all models. The best and second best results in each column are shown in \textbf{bold} and {\ul underline}.}
\label{table:style-consistency}
\end{table}

\section{Related Work}

\noindent\textbf{Personalized Dialogue Model} \quad Recent works on personalized response generation mainly fall into three categories: (1) those that personalize the response with a user-specific embedding learned during training  \cite{li-etal-2016-persona, chan-etal-2019-modeling}, (2) those that personalize the response with explicit user profiles or persona description sentences \cite{zhang-etal-2018-personalizing, DBLP:journals/corr/abs-1901-09672, ijcai2019p721, song-etal-2021-bob}, and (3) those that personalize the response with an implicit user persona extracted from user history conversations \cite{bak-oh-2019-variational, wu-etal-2021-personalized, 10.1145/3404835.3462828, zhong-etal-2022-less}. User-specific embeddings are shown to be ineffective and hard to generalize to unseen users since the embeddings need to be learned during training \cite{zhong-etal-2022-less}. Explicit user profiles and personas require manual data collection, which is very hard to scale up in practice and is often not available in deployed scenarios. Recent works \cite{10.1145/3404835.3462828, zhong-etal-2022-less} show strong scalability and robustness of the implicit user persona based method, and for that reason our work also focuses on this method.

The state-of-the-art implicit user persona method MSP \cite{zhong-etal-2022-less} incorporates a three-step hierarchical refiner to select informative tokens from relevant history responses from similar users, and the selected tokens are then prepended to the transformer deocder input as a prompt to personalize the generation process. However, their response selection module is trained on a news dataset that, because of domain divergence, may lead to sub-optimal retrieval performance on the intended dialogue task. Further, the hard discrete token selection module employed in MSP may be further improved by instead using a continuous prompt/prefix. Therefore, inspired by the hierarchical dialogue model and hierarchical transformer, we develop a personalized retrieval model that can use all user history conversations. As suggested by \citet{dudy-etal-2021-refocusing}, we develop a personalized generator with a prefix mechanism similar to that used in \citet{li-liang-2021-prefix} and \citet{liu-etal-2022-p}, but instead of learning the prefix during training, we train a prefix encoder to dynamically encode a personalized prefix with user history responses so that the model can be easily generalized to unseen users without further training.

Our model has two main differences from MSP: (1) we use a personalized response retriever trained on dialogue domain instead of a non-personalized retriever trained on distant news domain. (2) we use a dynamically encoded continuous prefix to fuse personalized retrieved responses to the generator rather than a discrete token prompt.

\noindent\textbf{Hierarchical Transformer} \quad Hierarchical transformers model long documents with a sentence-level transformer on top of a regular token-level transformer. 
The token-level transformer represents each sentence as a single vector embedding, and the embedding vectors of all sentences in the document are concatenated together and fed to the sentence-level transformer as input. 
These models are shown to be effective for long text classification \cite{DBLP:journals/corr/abs-1910-10781} and summarization \cite{zhang-etal-2019-hibert}. Our retrieval module uses a similar hierarchical transformer for response utterance-level embedding prediction, but differs in tasks and training strategy. Our retrieval module is trained on a generative next response prediction task with utterance-level causal masks.

\section{Conclusion}

In this work, we introduce RECAP, a personalized dialogue model, which generates responses in a retrieval augmentation manner. Unlike retrievers used in previous works, the hierarchical transformer retriever can perform personalized retrieval using user history responses. The context-aware encoder can encode and preserve the most useful information from the retrieved responses and fuse the information to a regular transformer decoder through continuous prefix vectors. Extensive experiments confirm that our model is capable of generating fluent, coherent, and personalized responses.

\section*{Ethical Issues}

Like most data-driven dialogue models, our model is trained on a large-scale naturally-occuring dataset, the Pushshift Reddit dataset \cite{DBLP:conf/icwsm/BaumgartnerZKSB20}, which may contain biased and offensive content.
To preserve persona and personal writing style as much as possible, we did not filter out conversations with this content.
To avoid potentially unethical responses in real-world usage, we suggest filtering out the data with unethical content before training or applying a post-generation filter for the offensive responses.

Even though our model is intended to generate personalized responses for only personal usage (e.g. personal virtual assistant), we realize it might be used for some malicious purpose by intentionally mimicking some individuals. 
Since our model is designed to be able to generalize to unseen users, we suggest keeping all personal data (i.e. personal dialogue history) local, as suggested by \citet{dudy-etal-2021-refocusing}, to minimize the risk of malicious imitation.
Retrieving from combined history conversations from multiple authors can potentially reduce the risk of exposing personal information of any specific user, but such use is not examined in this work.
Finally, we only allow the use of our model on public datasets or under the consent of the individuals being mimicked.

\section*{Limitations}

In this section, we discuss several limitations of our work that are worth future study.

First, the performance of the hierarchical transformer retriever is limited since the utterance-level transformer is trained from scratch only on our small-scale dataset due to limited time and computational resources.
With more resources, future work can focus on pre-training the utterance-level transformer on large-scale data such as the complete Pushshift Reddit data \cite{DBLP:conf/icwsm/BaumgartnerZKSB20}.
Pre-training can potentially improve the performance of the retriever and further improve the generation quality.

Second, the two types of retrieved responses in the RECAP-mixed model are encoded with the same encoder. 
However, intuitively, the two types of responses should contribute to generation in different ways, so treating them the same way might harm generation  performance.
This is also reflected in our results.
Even though the RECAP-mixed model shows improvement from both types of retrieved responses, the improvement is weaker than that on each separate model.
In future work, designing a split encoder for different types of retrieved responses may help maximally preserve the performance boost from both types of retrieved responses.

\section*{Acknowledgements}
This research is supported in part by the Office of the Director of National Intelligence (ODNI), Intelligence Advanced Research Projects Activity (IARPA), via the HIATUS Program contract \#2022-22072200006 and in part by DARPA (contracts \#HR001121C0169 and \#HR00112290025). The views and conclusions contained herein are those of the authors and should not be interpreted as necessarily representing the official policies, either expressed or implied, of ODNI, IARPA, or the U.S. Government. The U.S. Government is authorized to reproduce and distribute reprints for governmental purposes notwithstanding any copyright annotation therein. 
Approved for public release; distribution is unlimited. 

\bibliography{anthology,custom}
\bibliographystyle{acl_natbib}

\appendix

\section{Scientific Artifacts}
\label{sec:artifacts}

\subsection{Use of Existing Arifacts}

\begin{table}[!ht]
\begin{adjustbox}{max width=\columnwidth}
\begin{tabular}{lcc}
\toprule
Type                               & Name                      & License    \\ \midrule
\multirow{2}{*}{Dataset}           & Pushshift Reddit          & Not specified  \\
                                   & PANDORA                   & Not specified  \\ \midrule
\multirow{6}{*}{Pre-trained Model} & RoBERTa                   & MIT        \\
                                   & DialoGPT                  & MIT        \\
                                   & DHAP                      & Not specified        \\
                                   & MSP                       & Not specified  \\
                                   & Sentence-RoBERTa          & Apache-2.0 \\
                                   & Style-Embedding           & MIT        \\ \midrule 
\multirow{3}{*}{Library}           & HuggingFace Transformers & Apache-2.0 \\
                                   & Sentence Transformers     & Apache-2.0 \\
                                   & Scikit-learn              & BSD-3-Clause \\
\bottomrule
\end{tabular}
\end{adjustbox}
\caption{Licenses of artifacts used in this work.}
\label{table:artifacts}
\end{table}
The licenses of all scientific artifacts used in this paper is shown in \autoref{table:artifacts}. 
All artifacts with specified licenses are allowed to use in this work. 
The PANDORA dataset does not have a license, but we strictly follow their terms of use.\footnote{\url{https://psy.takelab.fer.hr/datasets/all/pandora/\#terms-of-use}}
All artifacts are intended to be used for research in machine learning and natural language processing, and our use is consistent with this intention.

\subsection{Created Arifacts}

We release a new model, RECAP in this work under the MIT license. Our model is only intended to be used personally or for research purposes. You can use it for yourself or on publicly available datasets. Using it to mimic other people without authorization is unethical and not allowed.

\section{T-test Details}
\label{sec:ttest}

\begin{table}[]
\small
\centering
\begin{adjustbox}{max width=\columnwidth}
\begin{tabular}{lcc}
\toprule
        & \# Subsets & Subset Size \\ \midrule
Reddit  & 100        & 2000        \\ \midrule
PANDORA \\
\hspace{2mm}-- Age/Gender   & 50         & 100         \\ 
\hspace{2mm}-- MBTI   & 50         & 500         \\
\midrule
Human   & 20         & 100         \\  
\bottomrule
\end{tabular}
\end{adjustbox}
\caption{T-test hyperparameters.}
\label{table:ttest}
\end{table}

For statistical significant tests, we randomly sample subsets from the test set and perform a paired t-test with the subsets' scores. The detailed hyperparameters are shown in \autoref{table:ttest}.

\section{More Experimental Results}
\label{sec:more}

\begin{table*}[]
\centering
\begin{adjustbox}{max width=\textwidth}
\begin{tabular}{@{\extracolsep{\fill}}lcccccccccc@{}}
\toprule
\multicolumn{1}{c}{\multirow{2.5}{*}{Model}} &                & \multicolumn{4}{c}{Token-overlap Metric}                        & \multicolumn{2}{c}{Learning-based Metric} & \multicolumn{2}{c}{Style Metric} \\ \cmidrule(rl){3-6} \cmidrule(rl){7-8} \cmidrule(rl){9-10} 
\multicolumn{1}{c}{}                       & PPL\textdownarrow            & BLEU-1\textuparrow         & BLEU-2\textuparrow        & ROUGE-L\textuparrow        & METEOR\textuparrow        & BERTScore\textuparrow           & BLEURT\textuparrow              & Embed Sim\textuparrow       & CAV Acc\textuparrow        \\ \midrule
DialoGPT                                   & 31.25$^\ddagger$          & 11.54$^\ddagger$          & 3.26$^\ddagger$          & 8.49$^\ddagger$           & 6.86$^\ddagger$          & 0.4240$^\ddagger$              & 0.2421$^\ddagger$              & 22.15$^\ddagger$           & 51.14$^\ddagger$          \\
DialoGPT + history                         & 29.66$^\ddagger$          & 12.09$^\ddagger$          & 3.89$^\ddagger$          & 9.84$^\ddagger$           & 7.82$^\ddagger$          & 0.4361$^\ddagger$              & 0.2482$^\ddagger$              & 40.66$^\ddagger$           & 64.02          \\
DHAP                                       & 29.99$^\ddagger$           & 14.09$^\ddagger$           & 4.37$^\ddagger$           & 9.73$^\ddagger$            & 7.86$^\ddagger$           & 0.4323$^\ddagger$               & 0.2511$^\ddagger$               & 37.14$^\ddagger$            & 61.53$^\ddagger$           \\
MSP                                        & 30.47$^\ddagger$           & 12.95$^\ddagger$           & 3.96$^\ddagger$           & 9.36$^\ddagger$            & 7.51$^\ddagger$           & 0.4307$^\ddagger$               & 0.2453$^\ddagger$               & 34.72$^\ddagger$            & 59.61$^\ddagger$           \\ \midrule
CAP                                        & \textbf{29.44} & 14.78$^\ddagger$          & 4.61$^\ddagger$           & 10.09$^\ddagger$           & 8.14$^\ddagger$           & 0.4356$^\ddagger$               & 0.2534$^\ddagger$               & 40.38$^\ddagger$            & 63.71$^\ddagger$          \\
RECAP-style                              & 29.54$^\ddagger$          & 14.79$^\ddagger$          & 4.61$^\ddagger$          & 10.05$^\ddagger$          & 8.08$^\ddagger$          & 0.4350$^\ddagger$              & 0.2525$^\ddagger$              & \textbf{41.40}  & {\ul 64.14}    \\
RECAP-semantic                           & 29.50$^\ddagger$          & \textbf{15.12} & \textbf{4.77} & \textbf{10.33} & \textbf{8.31} & \textbf{0.4617}     & \textbf{0.2749}     & 39.65$^\ddagger$           & 64.00          \\
RECAP-mixed                              & {\ul 29.47}$^\dagger$    & {\ul 15.06}$^\dagger$    & {\ul 4.71}$^\dagger$    & {\ul 10.27}$^\dagger$    & {\ul 8.27}$^\dagger$    & {\ul 0.4372}$^\dagger$        & {\ul 0.2557}$^\dagger$        & {\ul 40.74}$^\dagger$     & \textbf{64.17} \\ \bottomrule
\end{tabular}
\end{adjustbox}
\caption{The automatic evaluation results on the Reddit dataset with perplexity, token-overlap metrics, learning-based metrics, and style metrics. The best and second best results in each column are shown in \textbf{bold} and {\ul underline}. "$\dagger$" and "$\ddagger$" indicates statistically significant difference for $p<0.05$, between the best or the top two models, respectively, determined by t-test.}
\label{table:auto-results}
\end{table*}

\begin{table*}[]
\centering
\begin{adjustbox}{max width=\textwidth}
\begin{tabular}{lC{5em}C{5em}C{5em}C{5em}C{5em}C{5em}C{5em}}
\toprule
\multicolumn{1}{c}{\multirow{2.5}{*}{Model}} & \multicolumn{2}{c}{Demographic}   & \multicolumn{5}{c}{MBTI}                                                                \\ \cmidrule(rl){2-3} \cmidrule(rl){4-8}
\multicolumn{1}{c}{}                       & Age\textuparrow             & Gender\textuparrow          & I/E\textuparrow             & S/N\textuparrow             & T/F\textuparrow             & J/P\textuparrow             & Average\textuparrow         \\ \midrule
DialoGPT                                   & 0.0425$^\ddagger$          & 0.6103$^\ddagger$          & 0.5013$^\ddagger$          & 0.4982$^\ddagger$                & 0.5317$^\ddagger$          & 0.4655$^\ddagger$          & 0.4992$^\ddagger$          \\
DialoGPT + history                         & 0.1008$^\ddagger$          & 0.6763$^\ddagger$          & 0.4986$^\ddagger$          & {\ul \textbf{0.5146}} & 0.5561$^\ddagger$          & 0.4826$^\ddagger$          & 0.5130$^\ddagger$          \\
DHAP                                       & 0.0829$^\ddagger$          & 0.6653$^\ddagger$          & 0.4997$^\ddagger$          & 0.4872$^\ddagger$                & 0.5473$^\ddagger$         & {\ul 0.5135}$^\dagger$    & 0.5119$^\ddagger$          \\
MSP                                        & 0.1226$^\ddagger$          & 0.6816$^\ddagger$          & 0.4832$^\ddagger$          & 0.4924$^\ddagger$                & 0.5449$^\ddagger$          & 0.4870$^\ddagger$          & 0.5019$^\ddagger$          \\ \midrule
CAP                                        & \textbf{0.2051} & {\ul 0.7077}$^\dagger$    & 0.4998$^\ddagger$          & 0.4965$^\ddagger$                & 0.5635$^\dagger$          & 0.5070$^\dagger$          & 0.5167$^\dagger$          \\
RECAP-style                                & {\ul 0.1822}    & 0.7001$^\dagger$          & {\ul 0.5167}$^\dagger$    & {\ul \textbf{0.5146}} & 0.5562$^\ddagger$          & \textbf{0.5185} & {\ul 0.5265}    \\
RECAP-semantic                             & 0.1699$^\dagger$          & \textbf{0.7303} & \textbf{0.5279} & 0.5124                & {\ul 0.5676}    & 0.5025$^\ddagger$          & \textbf{0.5276} \\
RECAP-mixed                                & 0.1392$^\ddagger$          & 0.6962$^\dagger$          & 0.5154$^\dagger$          & 0.5045$^\ddagger$                & \textbf{0.5719} & 0.5049$^\ddagger$          & 0.5242          \\ \midrule \midrule
Ground-truth                               & 0.2617          & 0.7477          & 0.5149          & 0.5064                & 0.5705          & 0.5108          & 0.5257          \\ \bottomrule
\end{tabular}
\end{adjustbox}
\caption{The automatic personal traits evaluation results on the PANDORA dataset. The best and second best results in each column are shown in \textbf{bold} and {\ul underline}. "$\dagger$" and "$\ddagger$" indicates statistically significant difference for $p<0.05$, between the best or the top two models, respectively, determined by t-test.}
\label{table:traits-results}
\end{table*}

The full automatic evaluation results are shown here in \autoref{table:auto-results} and \autoref{table:traits-results}. 
Metrics within each category in \autoref{table:auto-results} are overall consistent with each other.
All token-overlap metrics and learning-based metrics are consistent with the human annotated fluency and coherency scores on the top two models.
The style metrics are consistent with the human annotated style score on the top two models (with different order for CAV accuracy).
\autoref{table:traits-results} shows the full personal traits evaluation results.
The metrics within each category are less consistent with each other, but the top two models on all metrics are always one of our four models, except for the MBTI J/P score.

\end{document}